\documentclass[letterpaper, 10 pt, conference]{ieeeconf}

\IEEEoverridecommandlockouts
\usepackage{cite}
\usepackage{amsmath,amssymb,amsfonts}
\usepackage{algorithmic}
\usepackage{graphicx}
\usepackage{textcomp}
\usepackage{xcolor}
\usepackage{booktabs}
\usepackage{footnote}
\usepackage{array}
\def\BibTeX{{\rm B\kern-.05em{\sc i\kern-.025em b}\kern-.08em
    T\kern-.1667em\lower.7ex\hbox{E}\kern-.125emX}}

\title{Beyond Kinematics: Benchmarking Simulation Fidelity \\ for Muscle-Driven Imitation Learning}
\newcommand{\code}[1]{\texttt{#1}}
\usepackage{xcolor}
\newcommand{\mm}[1]{\texttt{\textcolor{red!60!black}{MM#1}}}
\newcommand{\hmod}[1]{\texttt{\textcolor{blue!60!black}{H#1}}}

\newif\ifanonymous

\ifanonymous
    \author{Anonymous Authors}
\else
    \author{Ayah G. Ahmad$^\dagger$, Claire E. Borden$^\dagger$, and Maegan Tucker
    \thanks{$\dagger$ Denotes equal contribution}
    \thanks{This work is supported by the Georgia Institute of Technology.}
    \thanks{Authors are with the Dynamic Mobility Lab at Georgia Tech, Atlanta, U.S. \texttt{\{ayah, cborden9, mtucker\}@gatech.edu}}
    }
\fi

\begin{document}

\maketitle
\begin{abstract}
In this work, we conduct a systematic comparison of two state-of-the-art motion-imitation reinforcement learning (MIRL) pipelines, one built on SCONE/HyFyDy and one built on MuJoCo/MyoSim. HyFyDy emphasizes physiological realism through detailed musculotendon modeling, while MuJoCo prioritizes computational efficiency and scalable policy learning. While recent work has demonstrated that both pipelines reproduce human kinematics with high fidelity, it remains unclear if they accurately capture the underlying neuromuscular behavior that produced the movement. This limitation is particularly important for robotic assistive-device design and control, where outcome measures such as muscle activation patterns and metabolic cost are often used as optimization targets. To conduct a systematic comparison, our work compares both pipelines using a common set of human motion-capture and electromyography (EMG) measurements. 
The results find that while both pipelines produce similar kinematics with relative accuracy, the muscle activations from HyFyDy are more aligned with the experimental EMG, as supported by the average pooled (RMSE, $r$) values for muscle activations from HyFyDy and MuJoCo: (0.164, 0.4) and (0.344, 0.11), respectively. While we conclude that the more advanced physiological realism of HyFyDy currently makes it more suitable for musculoskeletal modeling, both require further development to bring physiological realism to GPU-parallelizable simulation environments and advance robotic assistive device design. 

\end{abstract}

\section{Introduction}

Robotic assistive devices, including exoskeletons and prostheses, have the potential to improve mobility and independence for individuals with motor impairments \cite{shepherd2026roadmap, gehlhar2023review}. Developing controllers that generalize across users, tasks, and environments, however, remains a persistent challenge \cite{Gehlhar2023}, \cite{Siviy2023}. Since human experiments are costly, time-intensive, and particularly burdensome for clinical populations, physiologically plausible musculoskeletal simulation has been proposed as a scalable alternative for accelerating controller development and device design \cite{slade2024human}. Recent advances in reinforcement learning (RL) have further increased interest in musculoskeletal simulation as a platform for training and evaluating control policies \cite{song2021deep, Simos2025KINESIS, denizdurduran2022optimum, choi2026musculoskeletal}. However, the extent to which current simulation pipelines accurately reproduce the underlying neuromuscular behavior of humans remains unclear. This limitation is particularly important for assistive robotics, where measures such as muscle activation patterns, muscle recruitment strategies, and metabolic cost are often used to evaluate device efficacy \cite{nuesslein2023comparing, Fleming2021, Han2021}. Consequently, before musculoskeletal simulation can be relied upon as a tool for assistive-device design, its ability to predict physiologically meaningful muscle-level outcomes must be benchmarked against experimental human data.

\begin{figure}
    \centering
    \includegraphics[width=1\linewidth]{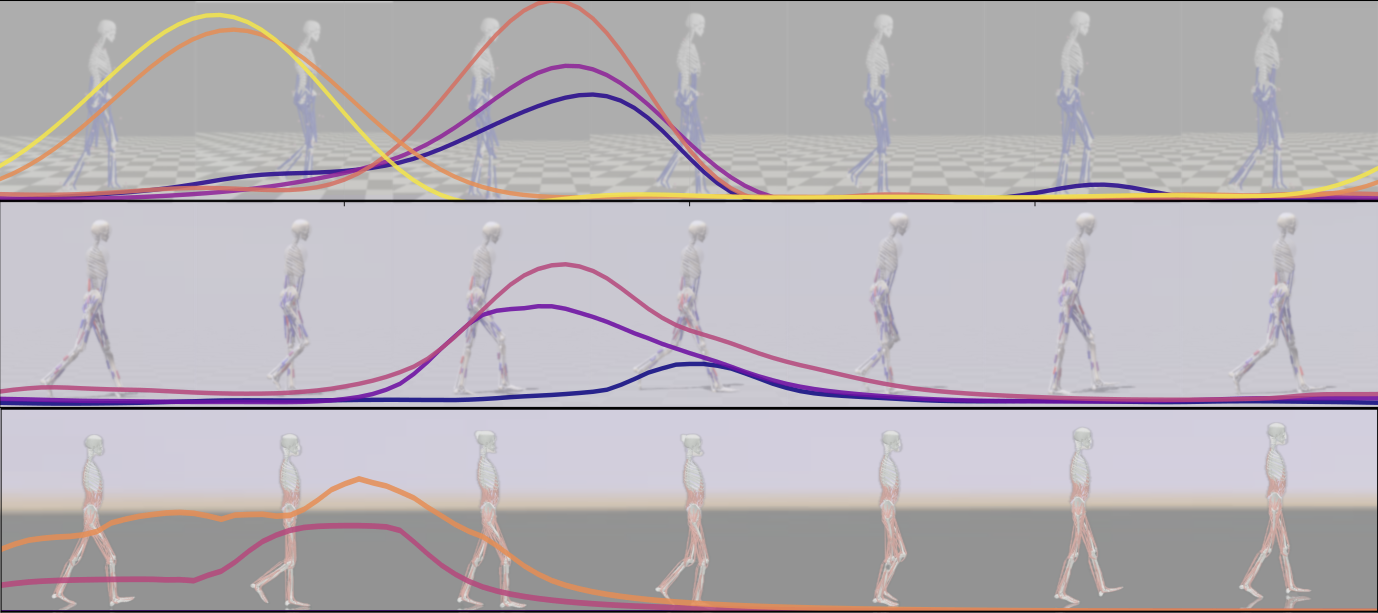} 
    \caption{This work benchmarks the predictive capabilities of two state-of-the-art musculoskeletal modeling simulation environments intended for motion-imitation-style reinforcement learning: HyFyDy/SCONE (middle) and MyoSim/MuJoCo (bottom), against real-world human data modeled in OpenSim (top). The gait tiles show each musculoskeletal model completing one gait cycle with the average speed-binned muscle activations for the medial gastrocnemius throughout it. Darker and lighter colors represent slower and faster speeds, respectively.}
    \label{fig: hero}
    \vspace{-5mm}
\end{figure}

\begin{figure*}[!t] 
    \label{fig: pipeline}
    \includegraphics[width=\textwidth]{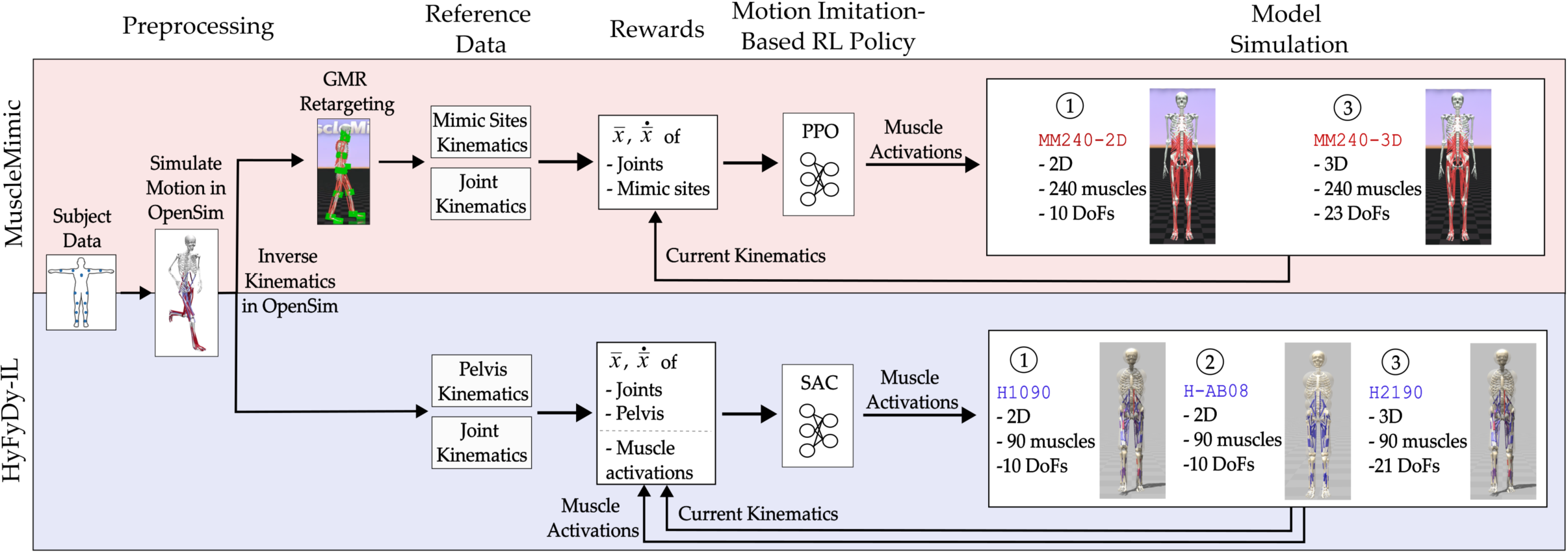} 
    \caption {Pipelines for MuscleMimic and HyFyDy-IL with modifications for this comparative study. The first models correspond to the comparison between the frameworks with similar standard models; the second model is the scaled HyFyDy model for the second comparison; and the third models are the 3D models for the final comparison.}
    \vspace{-5mm}
\end{figure*}

To address this gap, we benchmark two state-of-the-art musculoskeletal simulation pipelines that employ motion-imitation reinforcement learning (MIRL) to reproduce human locomotion from motion-capture demonstrations. In these approaches, human motion trajectories serve as reference signals for RL objectives, enabling the learned controller to reproduce observed movement while remaining physically consistent with the underlying musculoskeletal model. Our comparison moves beyond the evaluation of human kinematics alone and addresses whether these state-of-the-art pipelines are capable yet of accurately predicting the muscle activation patterns associated with those movements. 

The first pipeline we benchmark is the framework proposed by Choi et al. \cite{choi2026musculoskeletal}, which combines MIRL with the High-Fidelity Dynamics (HyFyDy) \cite{Geijtenbeek21} musculoskeletal simulator, featuring detailed musculotendon dynamics. The second is MuscleMimic \cite{musclemimic}, a MuJoCo-based \cite{todorov2012mujoco} MIRL framework designed for scalable training of large muscle-driven humanoid models. Both approaches are trained from human motion demonstrations and have demonstrated strong kinematic tracking performance. However, neither framework has been systematically validated against experimental electromyography (EMG) data, and no direct comparison currently exists between them.

In this work, we apply both MIRL frameworks in three comparative studies. First, a direct comparison of the two frameworks is performed using equivalent two-dimensional musculoskeletal models. Second, we examine the role of personalization by evaluating both generic and subject-specific musculoskeletal models. Lastly, the effect of model complexity is explored by comparing two-dimensional and three-dimensional implementations of each simulator. Together, these studies provide insight into how simulator fidelity, subject-specific modeling, and model dimensionality influence the physiological accuracy of MIRL. 

Our paper makes the following contributions:
\begin{enumerate}
    \item We introduce a benchmark for evaluating the muscle activation prediction accuracy of MIRL pipelines using synchronized motion-capture and surface EMG measurements. 
    \item We provide the first direct comparison, to our knowledge, of the HyFyDy and MuscleMimic musculoskeletal simulation environments for MIRL.
    \item We systematically evaluate the effects of simulator fidelity, subject-specific personalization, and model dimensionality on the prediction of muscle activation patterns during human locomotion.
    \item We provide an OpenSim-to-reference data for MuscleMimic and a trained subject-specific HyFyDy model to facilitate future benchmarking and reproducible evaluation of musculoskeletal simulation frameworks.
\end{enumerate}

\section{Background}

\subsection{Predictive Simulation}

Predictive simulation generates novel motions without necessitating the use of experimental data \cite{predictivemultibody}. In the context of robotic assistive devices, this can enable testing hypotheses, device designs, and treatment strategies before hardware validation. OpenSim \cite{opensim} is a dominant platform for physics-based musculoskeletal simulation, which contains Hill-type muscle-tendon dynamics, inverse kinematics and dynamics, and a full-body model library. Additionally, OpenSim can be used with the Neuromusculoskeletal Modeling  (NMSM) Pipeline to allow for patient-specific personalization \cite{nmsm}. Personalizing with NMSM, however, is time-consuming, taking over 25 hours to generate a model for an example patient \cite{nmsm}. For RL, these costs compound. Each new patient requires hours of offline fitting before training can begin, and training itself requires millions of environment interactions.

SCONE is an open-source program built for human and animal neuromuscular modeling that requires an additional musculoskeletal modeling software package, such as OpenSim \cite{opensim}, or HyFyDy \cite{Geijtenbeek2019}. HyFyDy includes the musculoskeletal models, packages, and scripts to simulate human motion, using a complete Millard musculotendon model with elastic tendons, variable pennation angles, and an error-controlled integrator running at approximately 7000 Hz \cite{Geijtenbeek21}.

Alternatively, MuJoCo is a different physics engine that is popular within the robotics community due to its high speed and soft contact models \cite{todorov2012mujoco}. MuJoCo is widely used for developing controllers for robots such as bipeds and quadrupeds \cite{EmbleyRiches2026, Wei2026SIMPLE}. It can also be paired with MyoSim \cite{Wang2022}, an open-source library with human musculoskeletal models. MuJoCo represents MyoSim's musculotendon units as a simplified Hill-type model, with completely inelastic tendons and a fixed timestep of 1000 Hz \cite{todorov2012mujoco}.

Both HyFyDy and MuJoCo have machine learning integration capabilities, making them suitable for training predictive musculoskeletal models. In our work, we compare HyFyDy and MuJoCo directly, using two state-of-the-art MIRL frameworks: for HyFyDy, we use an implementation inspired by Choi et al. \cite{choi2026musculoskeletal} (we term this framework HyFyDy-IL); for MuJoCo, we use the MuscleMimic framework \cite{musclemimic}. While both simulate human anatomy and locomotion, and have published impressive kinematic reference tracking, the degree to which they can predict human muscle activations patterns remains  unclear. This is the outcome we target in our comparisons.

\subsection{Reinforcement Learning Comparison}
Among the limited literature studying HyFyDy and MuJoCo in the context of predictive human modeling, Schumacher et al., compare the two simulators with the same RL framework \cite{schumacher2025emergence}. It was concluded that HyFyDy produces improved gait kinematics and ground reaction force predictions compared to MuJoCo when using models of similar complexity. The work also noted the limitations of both simulators in predicting muscle activation and their inability to produce accurate results, particularly for three-dimensional models with a similar muscle count. The inability of RL frameworks to accurately replicate muscle activations led some researchers to implement motion-imitation and attempt to model complex biological actuation in humans \cite{musclemimic, choi2026musculoskeletal}. Motion-imitation allows policies to move musculoskeletal models to mimic motion-capture data, decreasing the amount of necessary data to reproduce a motion and providing a method to potentially match poses and movements with muscle activations.  

\subsection{State-of-the-art RL Frameworks}

As previously mentioned, our work directly compares two state-of-the-art MIRL frameworks for musculoskeletal modeling: HyFyDy-IL and MuscleMimic \cite{choi2026musculoskeletal}, \cite{musclemimic}. Among the musculoskeletal modeling research, there exist other frameworks, such as Simos's KINESIS motion-imitation architecture \cite{Simos2025KINESIS}, yet these train for significantly longer times and increase computational expense drastically \cite{musclemimic}. HyFyDy-IL and MuscleMimic published kinematic and muscle activation results, showcasing their capabilities. However, additional research must be conducted to identify the most suitable simulator for musculoskeletal modeling, determine if MIRL is adequate for predicting human locomotion, and investigate limitations and potential advancements to improve these frameworks and simulators. The goal of our work is to contribute to the gap in comparative studies and provide further insight into musculoskeletal modeling frameworks and simulators. 

\begin{figure}
    \centering
    \includegraphics[width=0.62\linewidth,clip,trim=0 0 0 0]{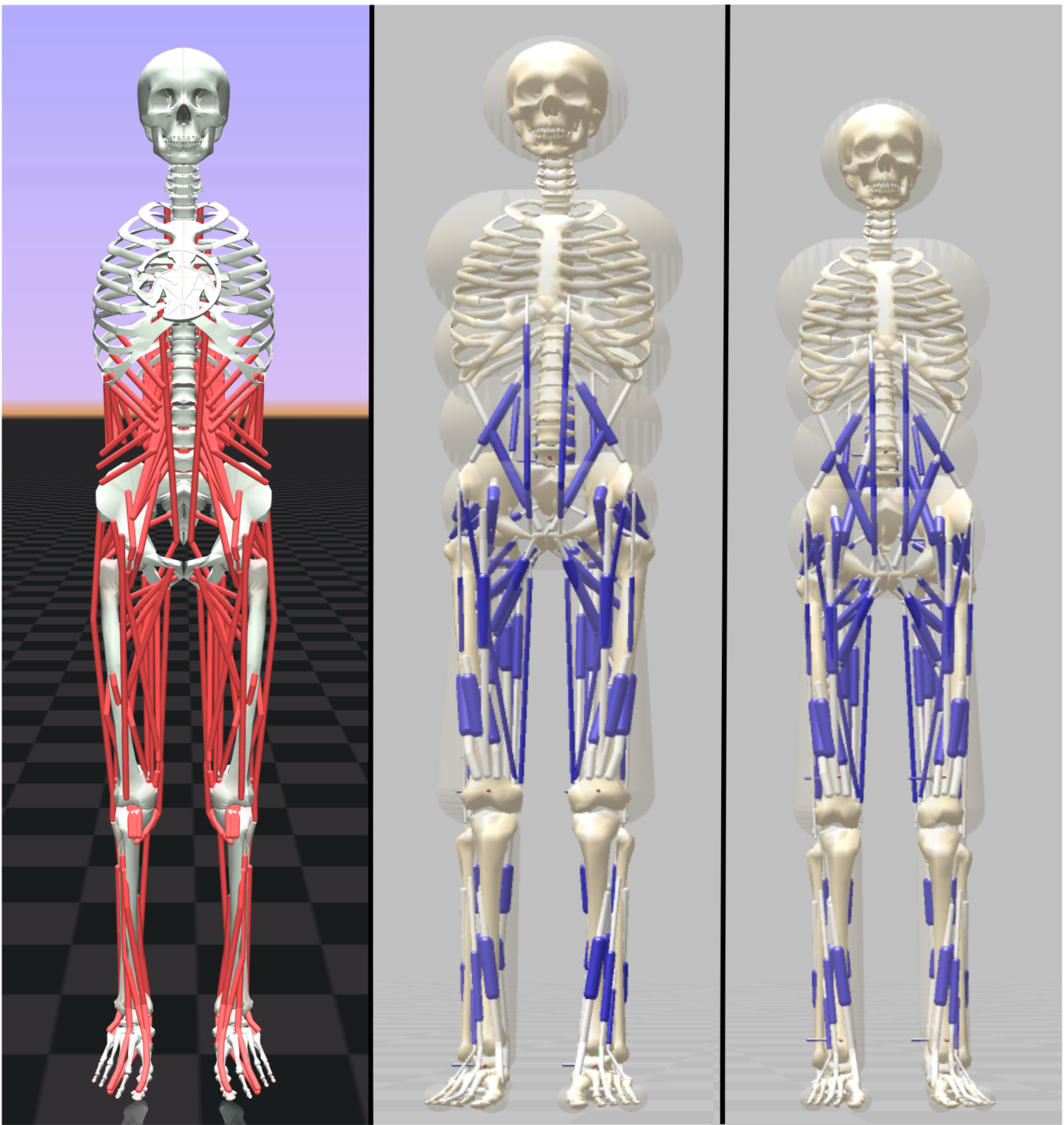}
    \caption{Two-dimensional models \mm{240-2D} (left), \hmod{1090} (middle), and \hmod{-AB08} (right). Not shown: \mm{240-3D} and \hmod{2190} (visually identical to \mm{240-2D} and \hmod{1090}).}
    \vspace{-6mm}
\end{figure}

\section{Systematic Comparison}

\subsection{Framework}
HyFyDy-IL is a CPU-based implementation that uses the SCONE physics engine, which incorporates Soft Actor-Critic (SAC) from Stable Baselines 3 \cite{sb3}. Due to its off-policy nature, SAC is highly sample efficient, making it particularly useful for learning with limited data \cite{sac}. Alternatively, MuscleMimic is a GPU-based JAX implementation that operates on the MuJoCo physics engine. MuscleMimic's policy includes a custom Proximal Policy Optimizer (PPO) that updates at each epoch, instead of after every certain number of epochs, to maintain on-policy learning and produce optimal results with parallel environment training \cite{musclemimic}, \cite{Schulman2017}. These underlying frameworks are kept constant to their originally published architectures to preserve consistency during our comparative study. 

\subsection{Preprocessing}
In the original architecture, MuscleMimic uses the AMASS \cite{Mahmood2019} and KIT Motion-Language \cite{Plappert2016} datasets. The AMASS dataset provides SMPL pose and body parameters, which are fit to 17 anatomical mimic sites on the MyoFullBody model. The motion can then be retargeted using two methods: 1) Mocap-Body, which uses MuJoCo's inverse kinematics, or 2) GMR-fit, which uses General Motion Retargeting's \cite{gmr} kinematic solver and applies joint and equality constraints to find joint configurations. 

Alternatively, HyFyDy-IL selects a reference walking motion obtained from the Scherpereel et al. open-sourced biomechanics dataset \cite{scherpereel2023dataset}, which includes five different walking speeds from 0.6 m/s to 2.2 m/s for approximately three minutes. The treadmill motion is converted to an overground walking motion and post-processed in OpenSim to compute corresponding joint angles using inverse kinematics with retained positions for body segments. These become the targets for MIRL.

To directly compare the two frameworks' ability to learn human locomotion, we train both HyFyDy-IL and MuscleMimic on the same data: a 140-second clip from subject AB08 in the Scherpereel et al. dataset \cite{scherpereel2023dataset}. For HyFyDy-IL, since the original framework uses a longer version of this data, no additional retargeting or preprocessing steps are performed. For MuscleMimic, however, the data is retargeted using GMR-fit to extract kinematic information for the mimic sites. This alignment of reference data preserves consistency in the preprocessing stage without affecting either policy's architecture and allows both simulators to be benchmarked against real-world patient data.

\subsection{Rewards}
Due to their unique architectures, HyFyDy-IL and MuscleMimic have different reward structures. HyFyDy-IL has a DeepMimic-style reward \cite{deepmimic, choi2026musculoskeletal}, focused on imitation tracking and effort penalties. The imitation rewards include joint position and velocity tracking and pelvic position tracking. Effort penalties impact the reward when 1) the muscle activations are too high and 2) their changes in activation are too significant, attempting to simulate real-life energy minimization while walking \cite{Selinger2015}. The total reward is:

\begin{small}
\begin{equation}
\begin{aligned}
R_{\mathrm{imit, H}} ={}&
w_{pos} R_{pos} + w_{vel} R_{vel}
+ w_{root} R_{root} \\& - w_{eff} C_{eff} - w_{\Delta eff} C_{\Delta eff}
\end{aligned}
\end{equation}
\begin{equation}
\begin{aligned}
R_{pos} =
\exp\left(
-\beta_{\mathrm{pos}}
\sum_i
\left(
\hat{\theta}_{i} - \theta_{i}
\right)^2
\right)
\end{aligned}
\end{equation}
\begin{equation}
R_{vel} =
\exp\left(
-\beta_{\mathrm{vel}}
\sum_i
\left(
\dot{\hat{\theta}} - \dot{\theta}_{i}
\right)^2
\right)
\end{equation}
\begin{equation}
R_{\mathrm{root}} =
\exp\left(
-\beta_{\mathrm{root}}
\left\|
\hat{p}_{\mathrm{root}} - p_{\mathrm{root}}
\right\|_2^2
\right)
\end{equation}
\begin{equation}
C_{\text{eff}} = \frac{1}{\mu_{eff}} 
\sum_{i=1}^Me_i^2
\end{equation}
\begin{equation}
C_{\Delta \text{eff}} = \frac{1}{\mu_{\Delta eff}} 
\sum_{i=1}^M (e_{i, curr}-e_{i, prev})^2
\end{equation}
\end{small}

The $\beta$ terms are temperature parameters, $M$ is the number of muscles, $\mu$ is the mean activation across all muscles, and the terms with a hat represent the reference motion targets, whereas those without are the predicted values from the simulation. Additionally, ``eff" refers to the muscle effort. 

MuscleMimic's rewards also have a DeepMimic-style \cite{musclemimic}. The rewards are described below, with $q$ and $\dot{q}$ as the respective joint angle and angular velocity; $p$, $\theta$, $\omega$, and $v$ as the respective mimic site position, orientation, angular velocity; and linear velocity, and $r$ and $\dot{r}$ are the respective root (pelvis) position and velocity. 

\begin{small}
\begin{equation}
\begin{aligned}
R_{\mathrm{imit, MM}} ={}&
w_q R_q + w_{\dot{q}} R_{\dot{q}}
+ w_p R_p + w_{\theta} R_{\theta} \\
&+ w_{\omega} R_{\omega}
+ w_v R_v 
\end{aligned}
\end{equation}
\begin{equation}
R_q = \exp\Bigg[
-\beta_q \Bigg(
\frac{1}{N}
\sum_{i=1}^{N}
\left(q_{i,lin}-q_{i,lin}^{*}\right)^2
+\bar{\theta}
\Bigg)
\Bigg]
\end{equation}
\begin{equation}
\bar{\theta}
=
\frac{1}{N_q}
\sum_{j=1}^{N_q}
\theta\left(
q_{j, quat},
q_{j, quat}^{*}
\right)
\end{equation}
\begin{equation}
R_r =
\exp\left[
-\beta_r
\frac{1}{N_r}
\sum_{i=1}^{N_r}
(r_i^{\mathrm{}}-r_i^{*\mathrm{}})^2
\right]
\end{equation}
\begin{equation}
R_{\dot{r}} =
\exp\left[
-\beta_{\dot{r}}
\frac{1}{N_{\dot{r}}}
\sum_{i=1}^{N_{\dot{r}}}
({\dot{r}}_i^{\mathrm{}}-{\dot{r}}_i^{*\mathrm{}})^2
\right]
\end{equation}
\begin{equation}
R_p =
\exp\left[
-\beta_p
\frac{1}{K-1}
\sum_{i=1}^{K-1}
\left\|
p_i^{\mathrm{}}-p_i^{*\mathrm{}}
\right\|_2
\right]
\end{equation}
\begin{equation}
R_{\theta} =
\exp\left[
-\beta_{\theta}
\frac{1}{K-1}
\sum_{i=1}^{K-1}
\left\|
\phi_i^{\mathrm{}}
-\phi_i^{*\mathrm{}}
\right\|_2
\right]
\end{equation}
\begin{equation}
R_{\omega} =
\exp\left[
-\beta_v
\frac{1}{K-1}
\sum_{i=1}^{K-1}
\left\|
\omega_i^{\mathrm{}}
-\omega_i^{*\mathrm{}}
\right\|_2
\right]
\end{equation}
\begin{equation}
R_v =
\exp\left[
-\beta_v
\frac{1}{K-1}
\sum_{i=1}^{K-1}
\left\|
v_i^{\mathrm{}}
-v_i^{*\mathrm{}}
\right\|_2
\right]
\end{equation}
\end{small}

Here, $N$ is the number of joints, $K$ is the number of mimic sites, $\phi$ represents the rotation as described in \cite{musclemimic}, $D$ is the dimension of the action space, and each quantity in the norms are computed relative to the reference frame (the pelvis). 

Despite their differences, both frameworks rely on imitating position and velocity and choose to weigh position more heavily than other factors, as shown in Table \ref{tab:reward_params}. The reward structures are maintained for our comparative study, as changing either one could lead to additional changes that would alter the architecture and affect performance. 

\begin{table}[htbp]
\centering
\caption{Original coefficients for HyFyDy-IL (top) and MuscleMimic (bottom) reward functions.}
\label{tab:reward_params}
\begin{tabular}{p{0.9cm} p{4cm} >{\centering\arraybackslash}p{0.9cm} >{\centering\arraybackslash}p{0.9cm}}
\toprule

Term $k$ & Tracked quantity & $\beta_k$ & $w_k$ \\
\midrule
$\mathrm{pos}$          & Kinematic pose error              & 1.0 & $0.60$ \\
$\mathrm{vel}$           & Kinematic velocity error      & 1.0  &  $0.15$ \\
$\mathrm{root}$           & End-effector position error     & 80.0 &  $0.25$ \\
$\mathrm{eff}$        & Muscle activation magnitude  & 1.0 & $-2.00$ \\
$\mathrm{\Delta eff}$ & Muscle excitation rate  & 1.0 & $-2.00$ \\
\midrule
\multicolumn{3}{l}{Total ($\sum_{\text{k}} |w_k|$)} & $5.00$ \\
\bottomrule

\toprule
Term $k$ & Tracked quantity & $\beta_k$ & $w_k$ \\
\midrule
$\mathrm{q}$            & Joint positions              & $10.0$  & $0.10$ \\
$\mathrm{\dot{q}}$            & Joint velocities             & $2.0$   & $0.10$ \\
$\mathrm{r}$            & Root coordinates $(x, z, \theta)$ & $10.0$  & $0.10$ \\
$\mathrm{\dot{r}}$         & Root twist                   & $10.0$  & $0.10$ \\
$\mathrm{p}$            & Relative site positions      & $100.0$ & $0.60$ \\
$\mathrm{\theta}$            & Relative site orientations   & $10.0$  & $0.01$ \\
$\mathrm{\omega}$   & Relative site angular vel.   & $0.1$   & $0.10$ \\
$\mathrm{v}$        & Relative site linear vel.    & $0.1$   & $0.10$ \\
\midrule
$\mathrm{Total (\sum_k |w_k|)}$ &  &  & $1.21$ \\
\bottomrule
\end{tabular}
\end{table}

\subsection{Musculoskeletal Models}

\subsubsection{Aligning Base Musculature and Bone Structure}
The original HyFyDy-IL framework uses the three-dimensional \hmod{2190} model, with 21 degrees of freedom, 90 muscles, and no bone structure for the arms \cite{choi2026musculoskeletal}, while MuscleMimic uses MyoFullBody, a model with a complete bone structure, 72 degrees of freedom, 123 joints, and 416 muscles \cite{musclemimic}.
Specific changes are made to align the musculature and bone structure of the two models. The models differ in how the individual muscles are subdivided (eg, for each leg, MyoFullBody represents the gluteus maximus as three separate actuators, \code{glmax1, glmax2, glmax3}, while \hmod{2190} uses a single muscle). Thus, instead of restricting the MyoFullBody model to 90 muscle actuators, the primary muscle groups are aligned. The resultant model \mm{240} thus has 240 muscles, representing the same muscles as those in \hmod{2190}. Additionally, for \mm{240}, the arm structures are removed, and the bones of the torso are fused together to represent and match HyFyDy's rigid torso in two dimensions.

\subsubsection{Comparison Setup}
We train five models:
\begin{enumerate}
    \item \textbf{\hmod{1090} (\textit{HyFyDy, 2D, 10 DoF, 90 muscle actuators)}: }The \hmod{2190} model constrained to two-dimensions by removing 11 degrees of freedom.
    \item \textbf{\hmod{-AB08} (\textit{HyFyDy, 2D, 10 DoF, 90 muscle actuators}): }The \hmod{1090} model scaled to match patient AB08's parameters from the Scherpereel et al. dataset \cite{scherpereel2023dataset} and represent the subject. It maintains the same number of muscles and DoFs, but the body geometric, mass, and inertia scale factors from the OpenSim model replace the default values in \hmod{1090}. The muscle actuators are unaltered as it is assumed that muscle size and force do not change significantly between subjects or models. Notably, we choose not to use HyFyDy's OpenSim-to-HyFyDy conversion tool, which does not include contact geometries needed for ground reaction forces or an architectural complexity similar to the other HyFyDy models (it contains 54 muscles). 
    \item \textbf{\hmod{2190} (\textit{HyFyDy, 3D, 21 DoF, 90 muscle actuators}): }The default 3D model provided by HyFyDy.
    \item \textbf{\mm{240-2D} (\textit{MuscleMimic, 2D, 10 DoF, 240 muscle actuators}):} The \mm{240} model constrained to two dimensions to align with \hmod{1090}.
    \item \textbf{\mm{240-3D} (\textit{MuscleMimic, 3D, 23 DoF, 240 muscle actuators}):} The \mm{240} model constrained to three dimensions. 
\end{enumerate}


\begin{table*}[htbp]
\centering
\caption{Resulting Joint Kinematics and Muscle Activations: $\mathrm{RMSE}$, $\mathrm{Pearson}$ $r$}
\label{tab:results}
\begin{tabular}{lcccccccccc}
\toprule
 & \multicolumn{2}{c}{\hmod{1090}} & \multicolumn{2}{c}{\hmod{-AB08}} & \multicolumn{2}{c}{\hmod{2190}} & \multicolumn{2}{c}{\mm{240-2D}} & \multicolumn{2}{c}{\mm{240-3D}\footnote{{\mm{240-3D} was excluded from statistical computations due to its inability to complete the 140-second walking motion, instead only completing approximately 18-seconds.}}} \\
\cmidrule(lr){2-3} \cmidrule(lr){4-5} \cmidrule(lr){6-7} \cmidrule(lr){8-9} \cmidrule(lr){10-11}
 & RMSE & $r$ & RMSE & $r$ & RMSE & $r$ & RMSE & $r$ & RMSE & $r$ \\
\midrule
\multicolumn{11}{l}{\textit{Joint Kinematics ($^{\circ}$)}} \\
Hip    & $11.26$          & $\mathbf{0.96}$ & $14.07$          & $0.93$          & $\mathbf{9.86}$  & $0.93$          & $11.04$          & $0.76$          & n/a          & n/a          \\
Knee   & $12.64$          & $\mathbf{0.96}$ & $21.76$          & $0.79$          & $\mathbf{12.42}$ & $0.87$          & $18.50$          & $0.70$          & n/a          & n/a          \\
Ankle  & $\mathbf{11.69}$          & $\mathbf{0.82}$ & $12.77$          & $0.57$          & $15.60$          & $0.53$          & $13.42$          & $-0.07$         & n/a &   n/a       \\
\midrule
Pooled & $\mathbf{11.88}$ & $\mathbf{0.89}$ & $16.68$          & $0.80$          & $12.85$          & $0.88$          & $14.65$          & $0.76$          & n/a         & n/a         \\
\midrule
\multicolumn{11}{l}{\textit{Muscle Activation (activation)}} \\
Tibialis Anterior     & $\mathbf{0.224}$          & $\mathbf{0.38}$ & $0.320$          & $0.08$          & $0.261$          & $0.33$          & $0.284$          & $0.36$          & n/a  & n/a          \\
Rectus Femoris     & $\mathbf{0.128}$ & $\mathbf{0.23}$          & $0.165$          & $-0.23$         & $0.209$          & $0.00$         & $0.630$          & $0.15$          & n/a         & n/a \\
Biceps Femoris     & $\mathbf{0.238}$ & $-0.27$         & $0.248$          & $-0.30$         & $0.254$          & $\mathbf{0.20}$         & $0.417$          & $0.23$          & n/a          & n/a \\
Gluteus Medius   & $\mathbf{0.085}$ & $\mathbf{0.53}$          & $0.097$          & $0.31$          & $0.129$          & $0.40$          & $0.187$          & -               & n/a          & n/a\\
Medial Gastrocnemius   & $\mathbf{0.131}$ & $\mathbf{0.81}$ & $0.221$          & $0.42$          & $0.278$          & $0.60$          & $0.232$          & $0.44$          & n/a          & n/a         \\
Vastus Lateralis     & $0.169$          & $\mathbf{0.61}$          & $\mathbf{0.077}$          & $-0.01$         & $0.261$          & $0.47$          & $0.177$          & -               & n/a & n/a \\
Adductor Longus   & $\mathbf{0.189}$          & $-0.08$         & $0.185$          & $\mathbf{-0.01}$         & $0.298$          & $-0.10$         & $0.366$          & $-0.38$         & n/a & n/a \\
Gluteus Maximus   & $\mathbf{0.052}$ & $\mathbf{0.44}$          & $0.083$          & $0.36$          & $0.187$          & $0.05$          & $0.201$          & -               & n/a          & n/a \\
\midrule
Pooled & $\mathbf{0.164}$ & $\mathbf{0.40}$ & $0.192$          & $0.20$          & $0.240$          & $0.26$          & $0.344$          & $0.11$          & n/a         & n/a         \\
\midrule
\end{tabular}
\end{table*}

\section{Results}
\begin{figure}[t] 
    \includegraphics[width=0.5\textwidth]{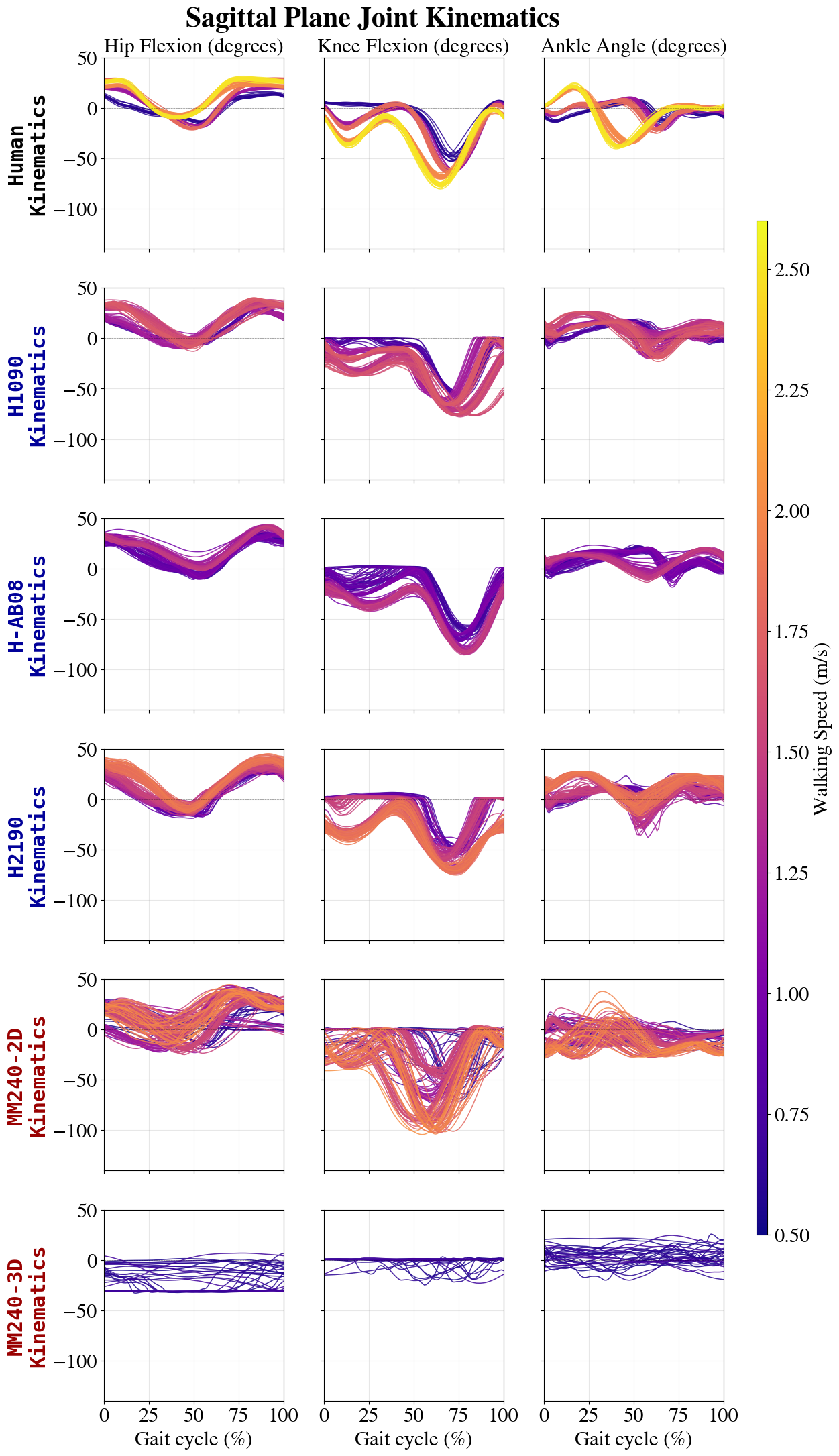} 
    \caption {Hip, knee, and ankle angle data from the human data (top) from the policies' rollouts over identical 140-second walking data, normalized over a gait cycle. Models \hmod{1090} (second row), \hmod{-AB08} (third row), and \hmod{2190} (bottom row), \mm{240-2D} (fourth row), \mm{240-3D} (fifth row).}
    \label{fig: all-kinematics}
    \vspace{-6mm}
\end{figure}

\begin{figure*}[t] 
    \includegraphics[width=\textwidth]{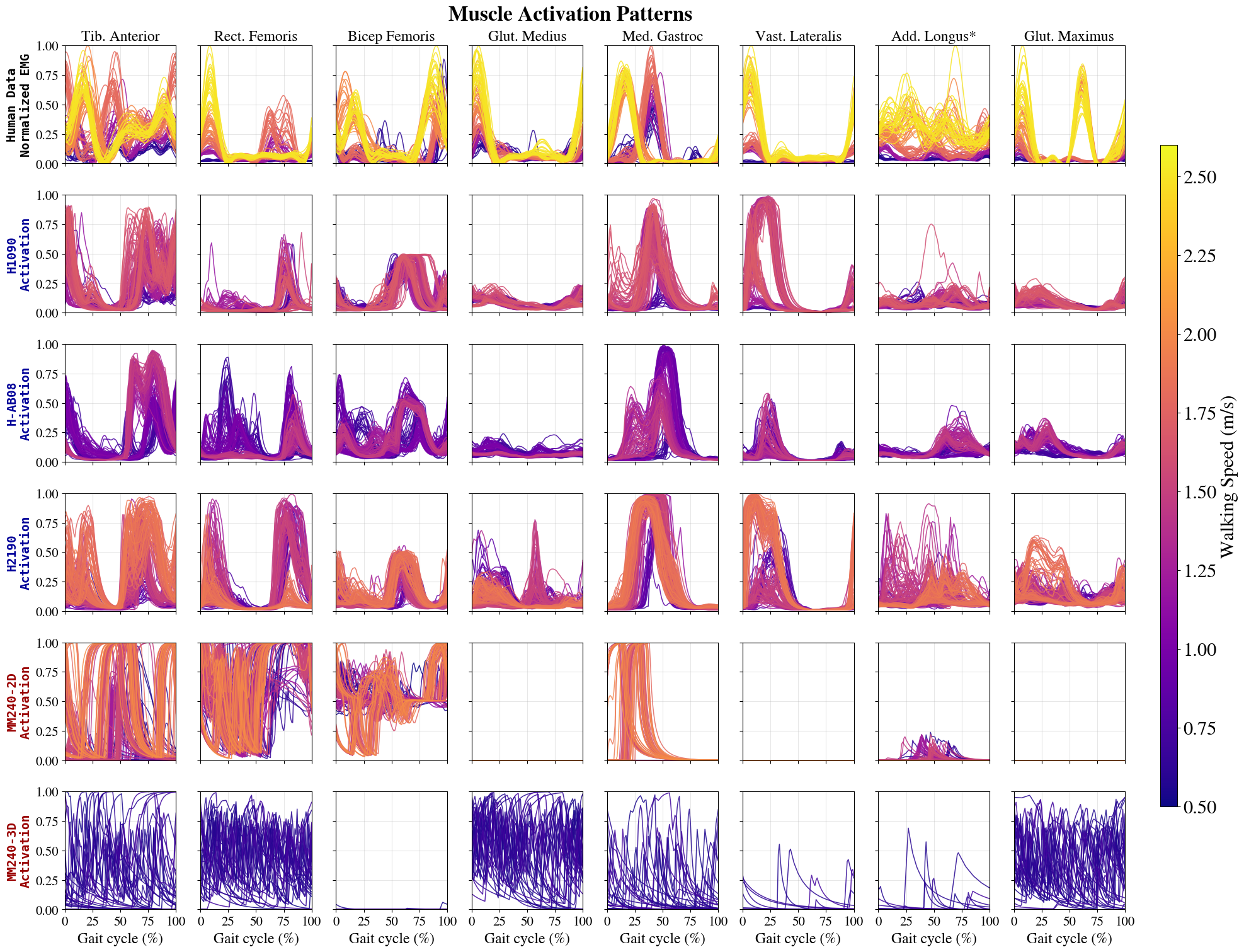} 
    \caption {The top row shows EMG experimental data recorded from subject AB08, which serves as the ground-truth benchmark for comparison. The rows below show muscle activations, normalized over a gait cycle, which are produced by each policy's rollouts on subject AB08's walking data. Eight muscles crucial for walking were analyzed: tibialis anterior, rectus femoris, biceps femoris, gluteus medius, medial gastrocnemius, vastus lateralis, adductor longus, and gluteus maximus. Models \hmod{1090} (second row), \hmod{-AB08} (third row), and \hmod{2190} (fourth row), \mm{240-2D} (fifth row), \mm{240-3D} (sixth row). Colors indicate the speeds the models were able to achieve. Note that none of the policies were able to reach the peak speed of 2.5 m/s. Also note that \mm{240-2D} learned not to activate several muscles: the gluteus medius, vastus lateralis, and gluteus maximus, which are necessary for walking.}
    \label{fig: all-activations}
\end{figure*}

Kinematic and muscle activation results are shown in Figures \ref{fig: all-kinematics} and \ref{fig: all-activations}, respectively. While the subject's data has speeds reaching over 2.0 m/s, none of the policies successfully reach these faster speeds. Eight major lower body muscles are used in our analysis, with plots of the additional muscles included in the appendix.

We compute two metrics for analyzing the prediction accuracy: Root Mean Squared Error (RMSE) and the Pearson $r$ correlation coefficient. RMSE quantifies the magnitude of disagreement of the model from the human; a lower RMSE indicates more agreement. Pearson $r$ quantifies shape agreement between the model and the human data; -1 indicates perfect negative correlation, 0 indicates no correlation, and 1 indicates perfect positive correlation. For each joint and muscle, RMSE and Pearson $r$ were computed between the human and model waveforms pooled across all gait-cycle points at matched speed bins.

\subsection{HyFyDy-IL versus MuscleMimic with Similar Models}
The first comparison evaluates the performance of \mm{240-2D} and \hmod{1090}, two-dimensional models similar to those used in the original MIRL frameworks. 
In terms of kinematics, both policies produce fairly accurate hip flexion results but similarly poor ankle flexion data. While their kinematic performance is similar, HyFyDy-IL produces lower RMSE values for six out of the eight recorded muscles and higher $r$ values for five. While HyFyDy-IL is able to replicate the EMG data better than MuscleMimic, neither produce highly accurate results, as \hmod{1090}'s pooled values are 0.164 and 0.40, and \mm{240-2D}'s are 0.312 and 0.16 for the RMSE and $r$ statistics, respectively. 

\subsection{Standard versus Personalized Models}
The second experiment explores how model personalization influences predictive accuracy by comparing the two-dimensional model from the first comparison, \hmod{1090}, with the scaled model, \hmod{-AB08}. MuscleMimic is excluded as available conversion tools, such as MyoConverter, are incapable of accurately scaling musculoskeletal models in MuJoCo while being compatible with MuscleMimic \cite{ikkala2022converting}. 

Both policies (\hmod{1090} and \hmod{-AB08}) show similar hip and knee angle trends, supported by relatively high $r$ values, with \hmod{1090} having slightly better results. Additionally, the \hmod{1090} produces a greater number of low-value RMSE and high-value $r$ results, such as for the medial gastrocnemius and rectus femoris muscles. The data demonstrates that the change in model kinematics alone may not be significant enough to influence the predictive capabilities of HyFyDy-IL. However, these results may change for subject models who are more dissimilar from the \hmod{1090} model.

\subsection{2D versus 3D Models}
The final study evaluates the effect of model complexity on the prediction accuracy by comparing the two- and three-dimensional models. Specifically, we compare \hmod{1090} with \hmod{2190} and \mm{240-2D} with \mm{240-3D}. Notably, the \mm{240-3D} and \hmod{2190} policies are not able to walk stably through the entire reference motion despite training for 36 hours on an NVIDIA A100 Tensorcore GPU and 32 hours on an AMD Ryzen Threadripper PRO 5945WX CPU (12 cores / 24 threads), respectively, indicating the increased difficulty of the training task.

For HyFyDy-IL, both policies (\hmod{1090} and \hmod{2190}) produce similar overall kinematics, with pooled RMSE and $r$ values of 11.88 and 0.89 for \hmod{1090} and 12.85 and 0.88 for \hmod{2190}, respectively. In terms of muscle activations, \hmod{1090} yields lower RMSE values for all muscles and higher $r$ values for seven of the eight muscle actuators compared to \hmod{2190}. Due to \mm{240-3D}'s inability to complete the walking motion for a significant portion of the reference data, it is excluded from the statistical results. Visually comparing the muscle activation plots \mm{240-2D} and \mm{240-2D}, it is clear that the three-dimensional model struggles to consistently actuate specific models. Overall, increased model complexity resulted in worse predictions across both frameworks.

\subsection{MuscleMimic with Modified Reward Structure}

Three similar MuscleMimic policies are trained to improve the first comparison by adding muscle activation penalties to align the reward function with HyFyDy-IL's. Since modifying the rewards themselves would significantly affect the structure of the MIRL framework, only the weight values are altered using three different methods. 
1) In HyFyDy-IL, the ratios of the rewards to the total sum of all $|w_k|$ is 20\%, while each muscle activation penalty weight makes up 40\% of the total weight, which are used to calculate the activation penalty terms in MuscleMimic. 2) In HyFyDy-IL, the ratios of the individual weights to the total sum of all $|w_k|$ are 12\% for ``position," 3\% for ``velocity," 5\% for ``root," and 40\% for both muscle activation penalties. MuscleMimic rewards are grouped into the same reward categories for HyFyDy-IL. The largest sum between the categories is maintained; other weight sums and their individual weights are scaled to match HyFyDy-IL's ratios, and the muscle activation penalties are calculated. 3) Weight values are chosen to be small (0.05 for each penalty term) and similar to the range of weights in the reward equation.

However, after training the three policies for 24 hours, the \mm{240-2D} fails to take even a single step. The poor results may be due to the calculated weights being too large, forcing the policy to learn more from activations than kinematics. While an ablation study would be valuable to tune these rewards, the long training time makes this infeasible without large amounts of compute. Although MuscleMimic produces undesirable muscle activations with a similar reward structure, the results do not signify its inability to train musculoskeletal models.  

\section{Conclusion}
Our comparative study evaluates the kinematic and muscle activation prediction capabilities of HyFyDy-IL and MuscleMimic with two-dimensional, personalized, and three-dimensional models. 
Each framework produces kinematics and muscle activations that best align with the experimental data using policies trained with their standard two-dimensional models, \hmod{1090} and \mm{240-2D}. The increase in dimensional complexity in the \hmod{2190} and \mm{240-3D} models led to higher pooled RMSE values and lower pooled Pearson $r$ values for both kinematics and muscle activations. Higher dimensionality leads the model to be more unstable during gait simulation, suggesting the need for longer training periods or implementation of termination conditions and reward functions that help stabilize the musculoskeletal model and produce symmetric motions in three-dimensional space. Additionally, the scaled two-dimensional model in HyFyDy (\hmod{-AB08}) also underperformed despite having the same complexity as \hmod{1090}. This may be due to small musculotendon unit displacements after scaling the bone structure of the \hmod{1090} model, affecting the muscle actuator's performance. 

Based on the overall results of the three comparison studies, HyFyDy-IL is currently more suitable for predicting human simulation. We hypothesize that this is due to its physiological realism. However, HyFyDy is not currently GPU-parallelizable, resulting in long training times. Thus, our research does not conclude that one framework or simulator is better than another. The work demonstrates that both frameworks and simulators are capable of simulating muscle-actuated human locomotion and require further development to improve their predictive capabilities.

To improve human locomotion simulation and prediction, this work can be extended to evaluate the simulation accuracy across individuals with diverse gaits. This would be beneficial for testing the robustness of the frameworks and provide insight into simulating the diversity seen in clinical populations. Along these lines, implementing assistive technology with the musculoskeletal model has significant potential and interest among the prosthetic and exoskeleton research community \cite{Zuo2026, Ryu2023}. Incorporating devices with accurate human simulation models could provide predicted human feedback to improve mechanical design and controller algorithms, advancing development without expensive, tiresome, and time-consuming physical prototyping and clinical testing \cite{Le2026, Mahmoudi2025}. Introducing robotic devices that interact with the musculoskeletal model would be a significant contribution to advance the biomechanics and assistive device engineering communities.














\section{Acknowledgements}
The authors would like to acknowledge and thank Ilseung Park and Inseung Kang for sharing their knowledge about HyFyDy and their assistance in setting up our environments, as well as the many researchers behind HyFyDy and MuscleMimic for developing the frameworks studied in this work. Models from Anthropic and OpenAI were used to assist in writing code for this project.  


\bibliographystyle{IEEEtran}
\vspace{-2pt}
\bibliography{references}
\end{document}